\documentclass[10pt,twocolumn,letterpaper]{article}

\usepackage[pagenumbers]{cvpr} 

\definecolor{cvprblue}{rgb}{0.21,0.49,0.74}
\usepackage[pagebackref,breaklinks,colorlinks,allcolors=cvprblue]{hyperref}
\usepackage{xcolor}

\def\confName{CVPR}
\def\confYear{2025}

\title{ProphetDWM: A Driving World Model for Rolling Out Future Actions and Videos}

\author{Xiaodong Wang$^{1,2}$  \quad Peixi Peng$^{1,2}$\thanks{Corresponding author.}   \\
 {$^{1}$Peking University \quad $^{2}$Peng Cheng Laboratory} \\
{\tt\small\{wangxiaodong21s@stu., pxpeng\}@pku.edu.cn}
}

\begin{document}
\maketitle
\begin{abstract}
Real-world driving requires people to observe the current environment, anticipate the future, and make appropriate driving decisions. This requirement is aligned well with the capabilities of world models, which understand the environment and predict the future. However, recent world models in autonomous driving are built explicitly, where they could predict the future by controllable driving video generation. We argue that driving world models should have two additional abilities: action control and action prediction. Following this line, previous methods are limited because they predict the video requires given actions of the same length as the video and ignore the dynamical action laws. To address these issues, we propose ProphetDWM, a novel end-to-end driving world model that jointly predicts future videos and actions. Our world model has an action module to learn latent action from the present to the future period by giving the action sequence and observations. And a diffusion-model-based transition module to learn the state distribution. The model is jointly trained by learning latent actions given finite states and predicting action and video. The joint learning connects the action dynamics and states and enables long-term future prediction. We evaluate our method in video generation and action prediction tasks on the Nuscenes dataset. Compared to the state-of-the-art methods, our method achieves the best video consistency and best action prediction accuracy, while also enabling high-quality long-term video and action generation.

\end{abstract}    
\section{Introduction}
\label{sec:intro}

World models are expected to understand the environment and predict the future conditioned on given actions \cite{lecun2022worldmodel,ha2018world}. 
Early works often model the world models in latent feature space and have been validated on games or simulators \cite{ebert2018visual,dosovitskiy2017carla}. Recently, inspired by the success of video generation ~\cite{openai2023sora}, several cutting-edge approaches \cite{hu2023gaia,wang2023driveWM,gao2024vista,yang2024genad} aim to build driving world models in an explicit way. That is, they could predict the future by controllable driving video generation and have shown promising results.




\begin{table*}[]
    \centering
    \begin{tabular}{lcccc}
    \toprule
    Method & Video Prediction & Action Control & Action Prediction & One-stage \\
    \midrule
    DriveGAN~\cite{kim2021drivegan}   & \checkmark &  &  \\
    DriveDreamer~\cite{wang2023drivedreamer} & \checkmark & \checkmark & \checkmark     & \\
    Drive-WM~\cite{wang2023driveWM} & \checkmark &  &      & \\
    ADriver-I~\cite{jia2023adriver} & \checkmark &  & \checkmark    & \\
    GenAD~\cite{yang2024genad} & \checkmark &  &      & \\
    Vista~\cite{gao2024vista} & \checkmark & \checkmark &     & \\
    \midrule
    Ours & \checkmark & \checkmark & \checkmark & \checkmark \\

    \bottomrule
    \end{tabular}
    \caption{Driving world model comparison. }
    \label{tab:dwm}
\end{table*}

\begin{figure*}
    \centering
    \includegraphics[width=\textwidth]{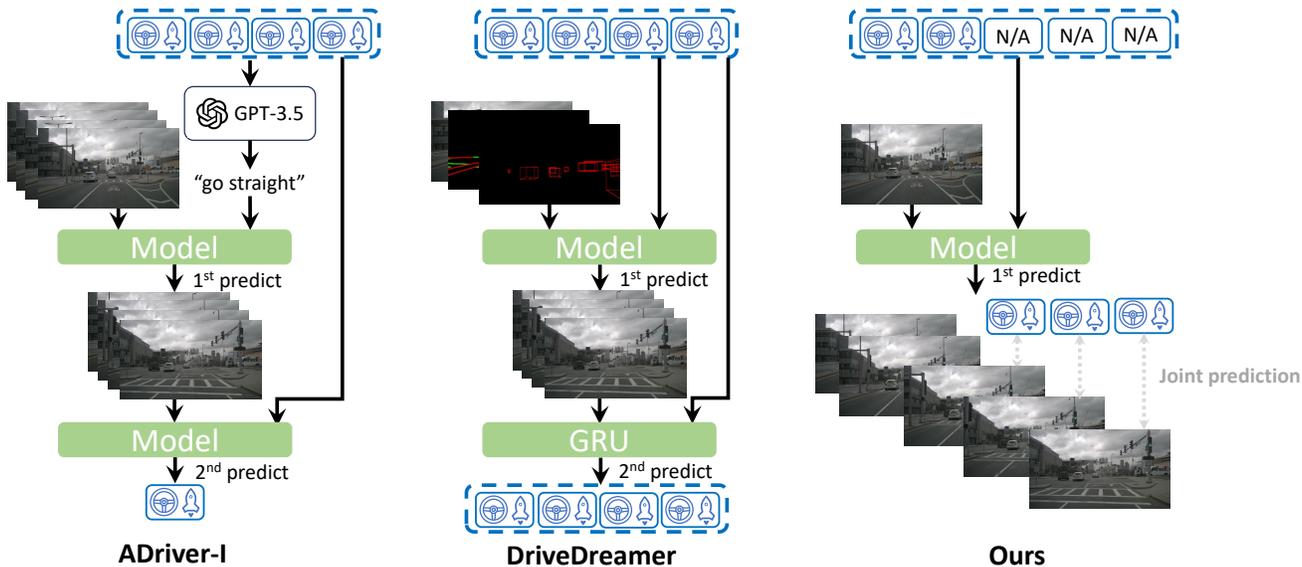}
    \caption{The comparison of video-action generation paradigms. We compare our method with previous methods that can perform action prediction. ADriver-I~\cite{jia2023adriver} and DriveDreamer~\cite{wang2023drivedreamer} belong to two-stage methods, because they need first predict the video, and then predict the subsequent actions, and their video prediction are conditioned on corresponding actions. In contrast, our method learns from the given action sequence and current observation and jointly predicts future actions and video in a one-stage manner. }
    \label{fig:intro}
\end{figure*}


Thanks to the development of diffusion models~\cite{ho2020ddpm,rombach2022sdm}, video prediction for driving scenarios is not very hard~\cite{blattmann2023align,kim2021drivegan,wang2023driveWM,jia2023adriver,yang2024genad}. Besides, we argue that the driving world model should have two additional abilities: 1) Action control: different from pure video generation, world models are more concerned with the counterfactual reasoning of actions. That is, the generated videos should exactly reflect the future differences of various given actions; 2) Action prediction: The world model is not only used to visualize the future as video generation methods do, but also plan the near future action simultaneously. The accurate action prediction is even more important in autonomous driving tasks. 
 Following this line, a comparison of recent driving world models is summarized in Table~\ref{tab:dwm}, where the action refers to low-level driving action (such as steering angle and speed). Specifically, models~\cite{gao2024vista,wang2023driveWM,kim2021drivegan} only focus on video generation, i.e., a kind of conditional prediction, given commands, trajectories, or low-level actions of every frame in the future, and ignore the action prediction. DriveDreamer~\cite{wang2023drivedreamer} and ADrive-I~\cite{jia2023adriver} are more similar to this work. However, they separate the predictions of video and action, and both predict video first and then use the predicted video to predict action. We argue this paradigm should be improved for two reasons:  Firstly, the length of the first video generation must be the same as the length of the given actions. Once the latter is limited, the long prediction must generated by multiple turns, and error accumulation may occur. Secondly, these methods only utilize the action sequence as a condition and ignore the dynamical action laws, which exhibit why such an action sequence is produced. Hence, as shown in Figure~\ref{fig:intro}, we propose a novel method that enables the driving world model to jointly learn to predict future actions and videos from given actions and observations.

To this end, we consider the controllable driving task to be a finite-horizon, partially observable Markov decision process (POMDP) \cite{smallwood1973pomdp}. Video in the driving process is considered to be observation, and low-level driving actions are considered to be actions performed by agents. To model the POMDP, we construct a driving world model named ProphetDWM that predicts future states and actions based on a sequence of given states and actions. Similar to most world models in latent feature space ~\cite{lin2023dynalang,hafner2023dreamerv3,hafner2020dreamerv2}, we also design two important components to learn the states representation and latent actions. Firstly, we propose an action module that aims to learn the latent action from the present moment to the future period. The action module is lightweight and only stacked with MLP (Multilayer Perceptron). It could not only learn the dynamics information from given actions but also incorporate the states of current observation to learn a sequence of latent actions, and the latent actions are decoded into low-level actions, which are supervised by the ground-truth actions. Second, we leverage the state-of-the-art diffusion models as a transition model to learn the state distribution, where the states are converted from image observations by a VAE (Variational Autoencoder) encoder. Besides the self-attention among the sequence states in diffusion models, we design a short pathway that injects the state context to the hidden states inside the diffusion model, i.e., the latent action extracted from the action module, which aims to promote the quality and consistency of predictions. Finally, the two components are trained jointly to predict the actions and videos.

To sum up, the major contributions of this paper are:
\begin{itemize}
    \item We propose ProphetDWM, an end-to-end driving world model that jointly learns to predict future actions and video. It is not a conditional low-level action-to-video model, but is trained by learning latent actions given finite states and predicting action and video.
    
    \item The world model has an action module to learn latent actions from the present moment to the future period by giving the action sequence and observations and a diffusion-model-based transition model to learn the state distribution. The world model is trained to predict the actions and videos jointly.

    \item We evaluate ProphetDWM in video generation and action prediction tasks on Nuscenes. Compared with state-of-the-art methods, our method achieves the best video consistency and action accuracy and can roll out high-quality long-term videos and actions.
\end{itemize}


\begin{figure*}
    \centering
    \includegraphics[width=\textwidth]{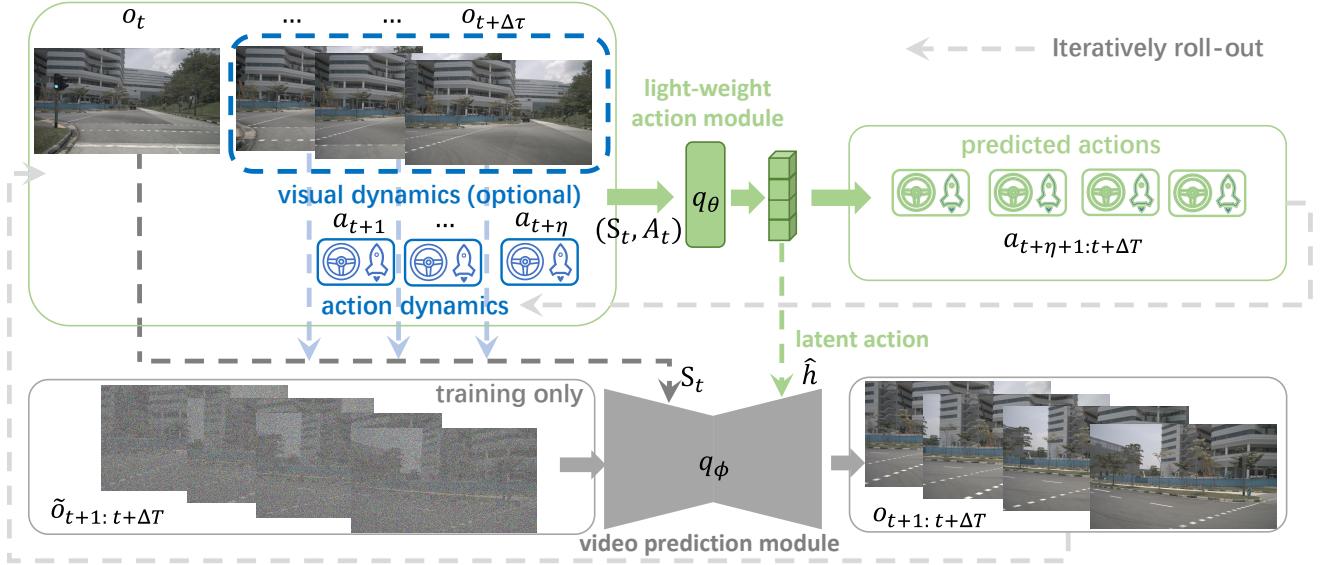}
    \caption{Schematic diagram of proposed ProphetDWM. Our world model has an action module to learn the latent action and a video prediction module to learn state distribution. All modules are optimized together to jointly predict future actions and videos.}
    \label{fig:main}
\end{figure*}
\section{Related Work}
\label{sec:relat}
\subsection{World Model}
The typical world model~\cite{lecun2022worldmodel} learns a general representation of the world and predicts future world states from history actions. Learning typical world models in the game~\cite{ha2018recurrent,hafner2019dreamer}, lab environments~\cite{ebert2018visual,tassa2018dmc}, or multimodal environments~\cite{lin2023learninglanguage} has been widely studied. Dreamer~\cite{wu2023daydreamer} and DreamerV2~\cite{hafner2020dreamerv2} learned a latent dynamic model to predict state values and actions in a latent space, which effectively handles the visual control tasks such as DeepMind Control Suite~\cite{tassa2018dmc}. MILE~\cite{hu2022mile} utilized imitation learning to learn a dynamic model and driving behavior in Carla~\cite{dosovitskiy2017carla}. Recently, a kind of definition of a world model is to purely predict the future~\cite{hu2023gaia,wang2024worlddreamer}. Moreover, driving world models in the real driving datasets have been explored~\cite{wang2023drivedreamer,jia2023adriver,wang2023driveWM,zhao2024drivedreamer2}. DriveDreamer~\cite{wang2023drivedreamer} trained a video diffusion model to receive conditions such as HDmap, 3Dbox, and actions to generate future driving videos. GAIA-1~\cite{hu2023gaia} designed a versatile diffusion model that learns image generation, video generation, and video interpolation. ADriver-I~\cite{jia2023adriver} leveraged a video diffusion model and a multi-modal language model that first generates a sequence of frames and then predicts the next action based on history action and all frames. Panacea~\cite{wen2023panacea} introduced BEV layouts guided video generation by integrating a 4D attention in U-Net. DriveWM~\cite{wang2023driveWM} took an innovative step towards generating multi-view videos. However, these previous works rely on future conditions to generate corresponding frames and failed to predict video and actions synchronously. In this work, we stepped towards joint action and video prediction.

\subsection{Video Generation}
Recent works in video generation focus on text-to-video generation, and the mainstream method is to pre-train a diffusion model on large-scale video-text datasets. Video-LDM~\cite{blattmann2023align} turned a pre-trained image LDM into a video generator by introducing temporal layers, which can also handle the generation of driving scenes. VideoPoet~\cite{kondratyuk2023videopoet} utilized a decoder-only transformer capable of processing multi-modal inputs, enabling zero-shot video generation. MagicVideo2~\cite{wang2024magicvideo} utilized a cascaded diffusion model that separates image and video generation. Lumiere~\cite{bar2024lumiere} designed a space-time U-Net that can directly generate the entire temporal duration of the video through a single pass in the model. MagicTime~\cite{yuan2024magictime} incorporated more physics knowledge and able to generate metamorphic videos. In this work, we designed a transition model based on the diffusion model that receives the learned prior from an action sequence and observations.
\section{Method}
\label{sec:method}
 \subsection{World Model Learning}
We consider the controllable driving task to be a finite-horizon, partially observable Markov decision process (POMDP~\cite{smallwood1973pomdp}). Visual observation, action space, and time horizon are denoted as $\mathcal{O}$, $\mathcal{A}$, and $\mathcal{T}$, respectively. \footnote{We don't consider reward here because it is correlated with specific task, and the reward often could be calculated by observations.} 
In this setting, the collected data in real-world autonomous driving include an agent (ego vehicle) that performs continuous actions $a_t$ on $o_t$ and receives $o_{t+1} \sim p(o_{t+1}|o_t, a_t)$.
In many realistic tasks such as driving, the POMDP could be observed only, and just $\{o_\tau, a_\tau\}_{\tau< t}$  is known. To evaluate the subsequent effects of different policies (i.e., action sequence), the world model is proposed to predict the future observations $o_{t+1:t+\Delta T}$ based on given action sequence $a_{t+1,t+\eta}$. In our setting, the length of the given action sequence is always much shorter than the length of required observation prediction ($\eta<\Delta T$), which is the main difference compared to previous methods \cite{wang2023drivedreamer,gao2024vista}. 







Since the goal of this model is expected to be a predictive task, and there are some benchmarks (such as Nuscenes~\cite{caesar2020nuscenes}, Waymo~\cite{waymo}) that collect multi-modal data from real-world autonomous driving environments, we can use supervised training and state-of-the-art models to model the distribution of the driving scenes and actions, and then easily to predict the future using the learned distribution. We choose the diffusion model~\cite{ho2020ddpm,rombach2022sdm} as the backbone to model the visual feature distribution. 

Our framework is shown in Figure~\ref{fig:main}. 
Given the image observation $o_t$ or historical observations $\{o_\tau\}_{\tau< t}$ ,  the latent state could be extracted by a VAE (Variational Autoencoder) encoder, and our model is trained on a sequence of latent features of videos, also, the predicted latent features can be decoded into an image by a VAE decoder. Note we only consider $o_t$ in our method, and it could be extended to $\{o_\tau\}_{\tau< t}$ directly. 
\begin{gather}
\begin{aligned}
&\text{VAE encoder:}  &&s_t \sim p_{\Phi}(s_t|o_t)
\\
& \text{VAE decoder:}  &&\hat{o}_t \sim p_{\Phi}(\hat{o}_t|s_t)
\label{eq:vae}
\end{aligned}
\end{gather}

The future observations  $o_{t+1:t+\Delta T}$ is closely related with the given future actions $a_{t+1:t+\Delta T}$. Considering $\eta<\Delta T$, a light-weight MLP as a multi-modal action sequence model as follows:

\begin{equation}
\text{Predicted action: } \tilde a_{t+\eta+1:t+\Delta T} \sim  q_{\theta}(o_{t},a_{t+1:t+\eta}).
\end{equation}
The action model not only predicts the future actions, but also extracts the latent action feature (See Sec.~\ref{sec:action}), which can be formulated as follows:
\begin{equation}
\text{Latent action feature: } \hat{h} \sim  q_{\theta}(o_{t}, a_{t+1:t+\eta})
\end{equation}

The predicted  $\tilde a_{t+\eta+1:t+\Delta T}$ provide the future actions for ego vehicle, while $\hat{h}$  could act as the condition to predict latent state $s_{t+1:t+\Delta T}$ because $\hat{h}$ contains the information of $\{a_{t+1:t+\eta}, \tilde a_{t+\eta+1:t+\Delta T}\}\}$:

\begin{equation}
\text{Transition model: }    q_{\phi}(\tilde{s}_{t+1:t+\Delta T} |s_t, \hat{h}).
\end{equation}
 Details are explained in Sec.~\ref{sec:video-trans}. After the transition model output a sequence of latent features $\tilde{s}_{t+1:t+\Delta T}$, these features are decoded into frames $\tilde{o}_{t+1:t+\Delta T}$ by a VAE decoder:

\begin{equation}
 \text{Predicted video: }  \tilde{o}_{t+1:t+\Delta T} \sim q_{\phi}(s_t).
\label{eq:predict}
\end{equation}

Besides, by iteratively rolling out the model prediction, we can simply obtain long-term future actions and videos.





\subsection{Light-weight Action Model}
\label{sec:action}
Instead of only predicting one action one time, which is low-efficiency, the action model here aims to predict multiple future actions with one step. Forecasting more future actions enables the application to become more flexible, and can specify the trade-off between speed and accuracy. 

The design motivation is that if the model can predict future actions well, then the high-dimensional features obtained during the prediction process have smaller losses than encoding the predicted actions again, which is also inefficient using re-encoding. Learning latent actions is
 also validated to be effective for predicting the future~\cite{bruce2024genie}

\begin{figure}
    \centering
    \includegraphics[width=\linewidth]{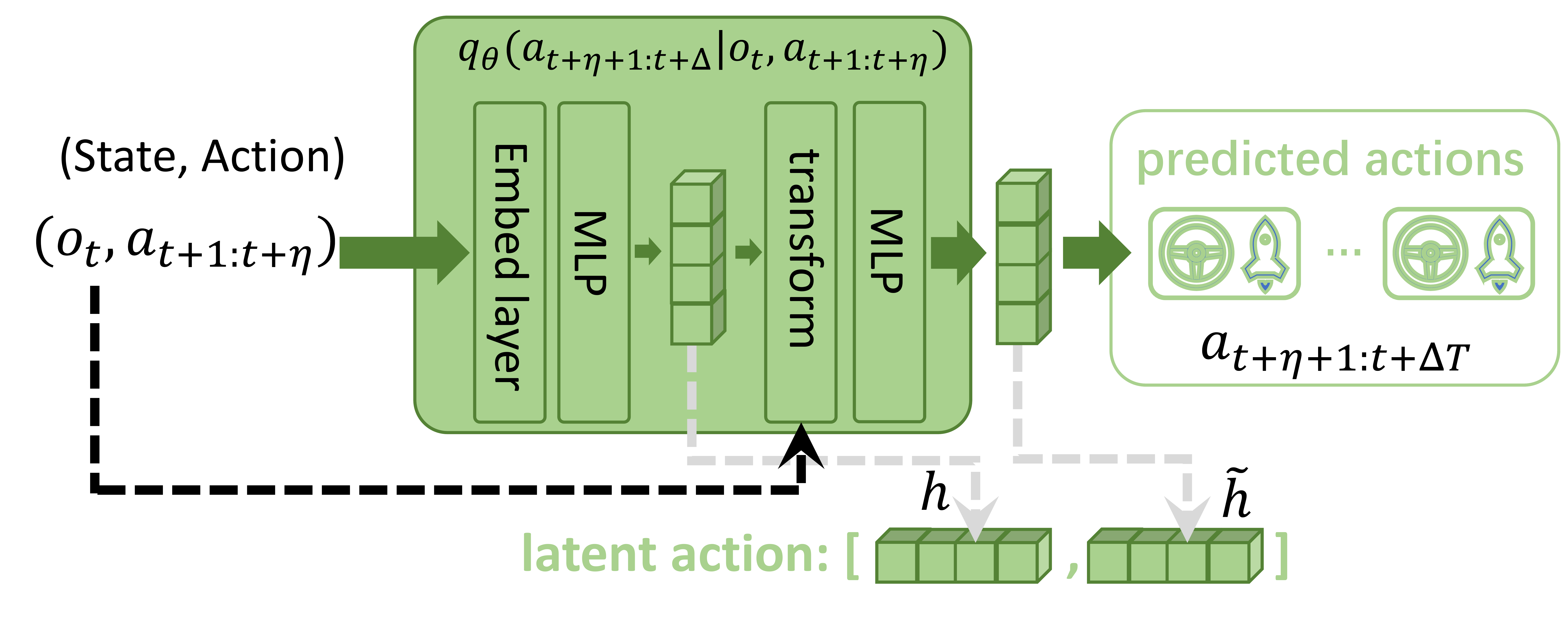}
    \caption{Light-weight action model. This model learns the latent action features given an action sequence and observations, }
    \label{fig:action}
\end{figure}

We carefully designed the action sequence model, the details are shown in Figure~\ref{fig:action}. Given actions $a\leq t+\eta $, the action model estimates the current action latent features and also the future action latent features by only feeding the known actions into the action model once:
\begin{equation}
h_{\leq t+\eta}, \Tilde{h}_{t+\eta+1:t+\Delta T} \sim q_{\theta}(a_{t+\eta+1:t+\Delta T}|o_t, a_{t+1:t+\eta})
\end{equation}
The action model receives multi-modal input including current states (observations) $o_t$ and known actions $a_{t+1:t+\eta}$. We use an embedding layer to lift the low-level actions to high-dimensional features and use an MLP to extract the temporal information, which can be regarded as the current latent action $h$. Then, a transform layer projects the visual features into action feature space, and the multi-modal features pass through an MLP to obtain the future latent action $\Tilde{h}$. We simply concatenate $h$ and $\Tilde{h}$ with the time dimension, resulting in final latent action $\hat{h}=[h, \Tilde{h}]$, which can prompt the following transition model to predict the consistency future frames. At the same time, the future latent action $\Tilde{h}_{t+\eta+1:t+\Delta T}$ are transformed into the future low-level actions $\Tilde{a}_{t+\eta+1:t+\Delta T}$. The action model is optimized by an L1 regularizer loss as follows:
\begin{equation}
    \min_{\theta} \mathcal{L}_a = \beta_a \| a_{t+\eta+1:t+\Delta T} - \Tilde{a}_{t+\eta+1:t+\Delta T} \|
\end{equation}

\subsection{Video Transition Model}
\label{sec:video-trans}
The world model learns rules from historical observations and action sequences to predict the future. We divide this process into two steps. The first step is written in Section~\ref{sec:action}, which is to learn the rules of action transition and obtain high-dimensional latent actions. For predicting the future, we can use state-of-the-art generative models, such as diffusion models~\cite{ho2020ddpm,rombach2022sdm}. We only need to design this model to receive latent actions and current observations. Therefore, in this section, we will introduce how to design and train a video transition model.

Given current observation $o_t$ and the latent action $\hat{h}$, we need to design a transition model that can predict future states. Intuitively, we can leverage diffusion models to achieve this purpose, as diffusion models are widely used in video generation~\cite{yuan2024magictime,wang2024magicvideo,yin2023nuwaxl} and prediction~\cite{yang2024cogvideox,ren2024consisti2v,blattmann2023svd}. However, most pre-trained diffusion models are conditioned on text, such as the text describing the scene, or depicting the dynamics and motion, where our target dataset Nuscenes~\cite{caesar2020nuscenes} is lacking. To this end, we utilized a state-of-the-art multi-modal language model (LLaVA-NeXT-Video~\cite{zhang2024llavanextvideo}) to annotate scenes in Nuscenes.

Based on annotations, we first fine-tuned a base model that adapts the general model~\cite{rombach2022sdm} to the driving scenario. Then, we design a spatiotemporal model following previous work~\cite{blattmann2023align,blattmann2023svd}. As shown in Figure~\ref{fig:main}, the model is built on the U-Net architecture. In order to satisfy the training of the U-Net and diffusion process, that is, the input and output feature dimensions are the same, and to learn from noise. We add noise to the state $s_t=p_\Phi(o_t)$ and ground-truth future states $s_{t+1:t+\Delta T}$, if we make a $\Delta T$-step prediction. The noisy state sequence is:
\begin{equation}
 \Tilde{s}_{t:t+\Delta T} = \sqrt{\overline{\alpha}_\tau}s_{t:t+\Delta T}+ (1-\overline{\alpha}_\tau)\epsilon, \  \epsilon \sim \mathcal{N}(\textbf{0}, \textbf{I})
\label{eq:noise_state}
\end{equation}
where $\alpha_\tau$ and $\overline{\alpha}_\tau$ are predefined hyper-parameters, $\tau$ denotes the noise timestep.
\paragraph{State context short pathway}To satisfy observations of varying lengths, because in general, more observations make more precise predictions, we designed a flexible fusion module to make better use of the observations $o_{t: (t+\Delta \tau)}$, where $\Delta \tau\ \geq 0$ is the length of additional observations. First, $o_{t: (t+\Delta \tau)}$ is fed into VAE to obtain latent states $s_{t: (t+\Delta \tau)\sim P_\Phi(o_{t: (t+\Delta \tau)})}$, then we repeat the last state to obtain the state context:
\begin{equation}
    s_c = [\underbrace{s_{t},s_{t+\Delta \tau}, \cdots, s_{t+\Delta \tau}}_{\Delta T \  \text{terms}}]
\end{equation}

Since subsequent observations are transitioned from the last observation $s_t$, the key of the fusion module is to provide a short pathway for the frame transition. The fusion module is designed with a multi-scale architecture that satisfies the multi-scale hidden states of the U-Net. 

Specifically, as for a short pathway at $d$-th block in the U-Net, the state context at $d$-th block $s_c^d$ is transferred to scale $w^d_c$ and shift $b^d_c$ via zero-initialized convolution layers. The scale $w^d_c$ and shift $b^d_c$ are fused into the hidden state $h^d$ of $d$-th block as follows:
\begin{equation}
    h^d= w^d_c \cdot h^d + b^d_c + h^d
\end{equation}
To support the classifier-free guidance sampling, we randomly drop the state context by a certain ratio in the training process. 
\begin{figure*}
    \centering
    \includegraphics[width=\linewidth]{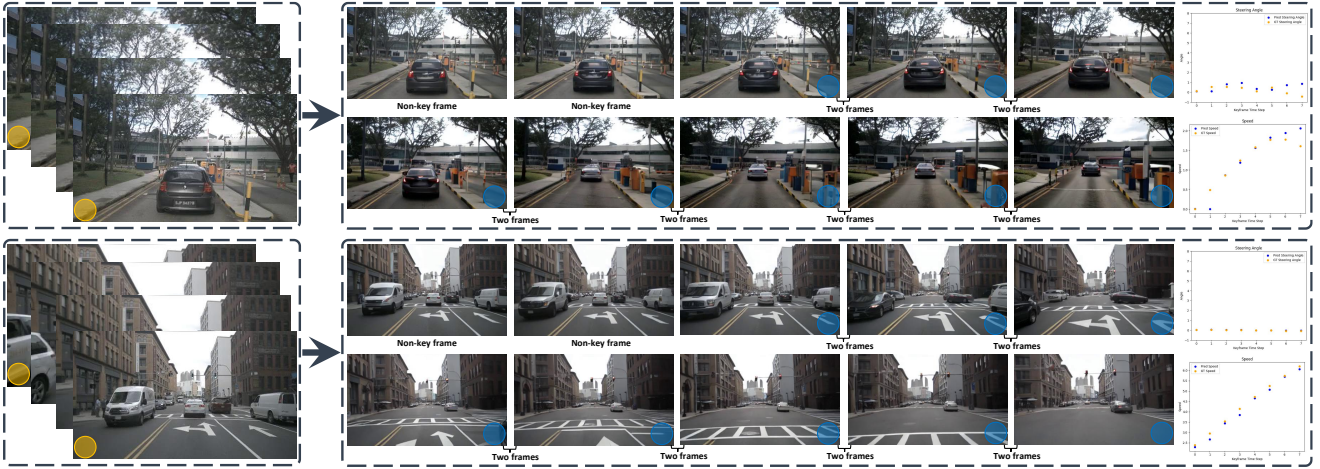}
    \caption{World model prediction by our proposed method. (Resolution of video is 256$\times$448. The orange \textcolor{orange}{\textcircled{ }} indicates the key frame with Ground-truth actions. The blue \textcolor{blue}{\textcircled{ }} indicates the key frame with the predicted actions, and the rest are non-key frames.) }
    \label{fig:re1}
\end{figure*}
\begin{figure*}
    \centering
    \includegraphics[width=\linewidth]{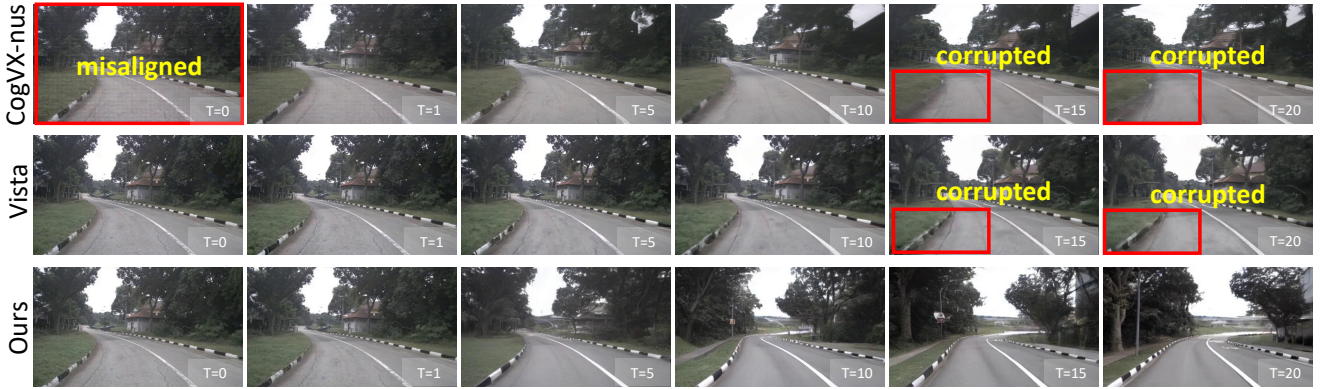}
    \caption{Video prediction comparison. Although different models have different resolutions for inference, CogVideoX-2b-nus(CogVX-nus) uses 480$\times$720, Vista uses 576$\times$1024, our model uses 256$\times$448, our model shows better quality and clarity, avoiding misaligned starting frame and corrupted results. }
    \label{fig:re2}
\end{figure*}

\paragraph{Future prediction}The video transition model learns to predict future frames. The action model in Sec.~\ref{sec:action} produces latent action $\hat{h}$ from the current observations $o_t$ and the current actions $a\leq t+\eta$, which provides the estimated dynamics and benefits the future frame prediction. The future prediction objective is:
\begin{equation}
    \min_{\phi} \mathcal{L}_v = \beta_v \mathbf{E}_{\tau,s,\epsilon} \| \epsilon - q_{\phi}(\Tilde{s}_{t:t+\Delta T}, s_c, \hat{h}, \tau) \|^2
\end{equation}

When training, the noise is added to known states $s_t$ and ground-truth states $s_{t:t+\Delta T}$, and this diffusion objective learns to predict the future states from the noise. Using the cross-attention mechanism, latent action $\hat{h}$ acts on the time dimension of the state sequence.

The world model is trained to optimize the overall loss $\mathcal{L}_a+\mathcal{L}_v$ with respect to all parameters.

\section{Experiment}

\begin{figure*}
    \centering
    \includegraphics[width=\linewidth]{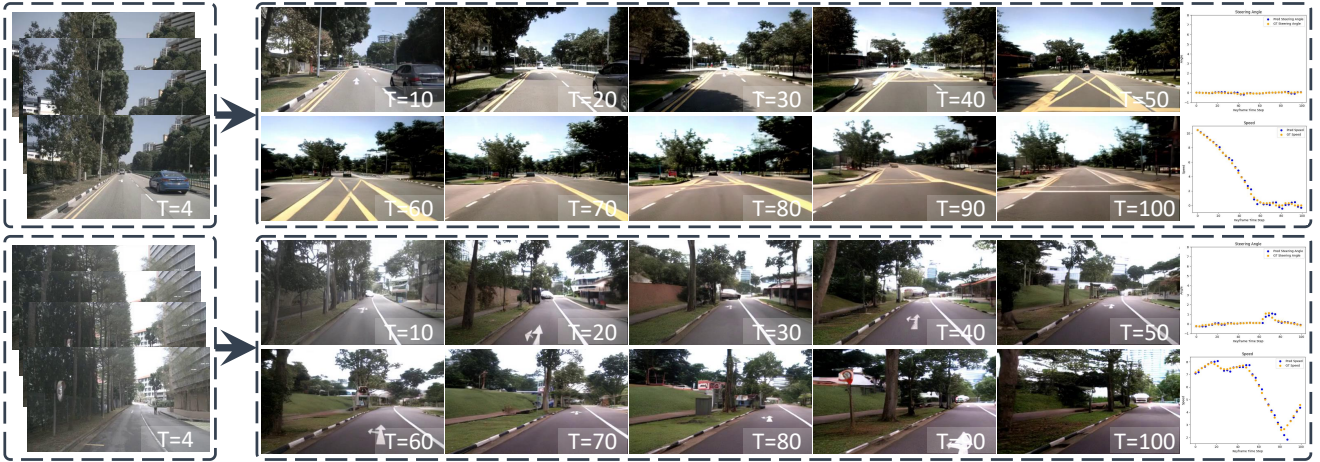}
    \caption{Long-term video and action rollout. Given four input frames, our model can predict long-term video and actions, up to 100 frames. The video quality is still well-maintained and the predicted actions are accurate. The upper sample shows that our model generates a future in which the ego car is slowing down and then slowly stopping. The bottom sample shows a future in which the ego car stays driving, then deviates slightly to the left (consistent with predicted steering angles), and then returns to a straight line.}
    \label{fig:long}
\end{figure*}

\begin{table*}[]
    \centering
    \caption{Quantitative comparison on driving video generation}
    \label{tab:fvd}
    \begin{tabular}{lccc}
    \toprule
    Method & Condition  & FID↓   & FVD↓    \\
    \hline
    DriveGAN~\cite{kim2021drivegan} & 1st image  &     73.4    & 502.3   \\
    DriveDreamer~\cite{wang2023drivedreamer} & 1st image+HDmap+3Dbox  &  52.6   & 452.0   \\
    DrivingDiffusion~\cite{li2023drivingdiffusion} & HDmap+3Dbox  & 15.9    & 335.0 \\
    Panacea~\cite{wen2023panacea} & 1st image+HDmap+3Dbox  & 20.4    & 247.0 \\
    GenAD-nus~\cite{yang2024genad} & 1st image & \underline{15.4} & 244.0 \\
    Drive-WM~\cite{wang2023driveWM} & 1st image+HDmap+3Dbox & 23.3 & \underline{228.5} \\
     \hline
    Ours & 1st image & \textbf{6.9} &  \textbf{190.5} \\
    \bottomrule
    \end{tabular}
    \vspace{-0.5cm}
\end{table*}

\begin{table*}[t]
   \caption{Quantitative comparison on driving action prediction. $^\dagger$ denotes action-only.}
  \label{tab:action}
  \centering
\begin{tabular}{lccc}
\toprule
    Method & L1↓ Speed   & L1↓ Steering angle & Average L1↓      \\ 
    \midrule
    DriverDreamer$^\dagger$ \cite{wang2023drivedreamer}& 0.150           & - & - \\
    DriverDreamer\cite{wang2023drivedreamer}          & 0.140           & - & - \\
    ADriver-I$^\dagger$ ~\cite{jia2023adriver}     & 0.122           & 0.101 & 0.1115 \\
    ADriver-I ~\cite{jia2023adriver}             & 0.103 & \textbf{0.092} & 0.0975  \\
    \midrule
    Ours & \textbf{0.090} & 0.093 & \textbf{0.0915} \\
    \bottomrule
  \end{tabular}
\end{table*}

\subsection{Setting}
All experiments are conducted on the Nuscenes dataset~\cite{caesar2020nuscenes}. Following the setting in~\cite{wang2023driveWM}, to make a fair comparison, we use the 192$\times$384 resolution for video quality evaluation. We use FID (Fréchet Inception Distance)~\cite{heusel2017fid} and  FVD (Fréchet Video Distance)~\cite{unterthiner2018fvd} to evaluate images and video quality, respectively. Besides, for video visualization and long-horizon rollouts, we train a 256$\times$448 video model. For action prediction, we follow~\cite{jia2023adriver} to report the L1 error of driving speed and steering angle. More details can be found in the Appendix.

\subsection{Qualitative Results}
\paragraph{Joint video-action prediction} Since we using joint training combining action prediction and future prediction, our trained model is capable of joint prediction, where predicted videos and actions are well aligned. We show two examples in Figure~\ref{fig:re1}. For each scene, we input four reference frames and two actions to prompt the dynamic information and then roll out future frames and actions iteratively. We present the first two non-key frames to indicate that our model can maintain the scenes well, and subsequent frames marked with blue circles are keyframes. The jointly generated actions are shown in the last column. For the upper sampler, the predicted future is a slow acceleration of the ego car. The bottom sampler presents a future with a faster acceleration. The predicted frames and actions are well aligned, validating the joint-prediction ability inside our world model.

\paragraph{Video quality comparison}
As there are limited open-sourced pre-trained models, DriverDreamer~\cite{wang2023drivedreamer}, ADrive-I~\cite{jia2023adriver}, and Drive-WM~\cite{wang2023driveWM} did not release the pre-trained models. We compare our method with Vista~\cite{gao2024vista}, the latest video prediction model in the driving scenario. We also fine-tune an open-source state-of-the-art model CogVideoX-2b~\cite{yang2024cogvideox} on Nuscenes as a video prediction model to be compared, denoted as CogVX-2b-nus. Different methods use different resolutions in the inference process.  CogVX-2b-nus uses 480$\times$720, Vista uses 576$\times$1024, and our model uses 256$\times$448 resolution. An example is shown in Figure~\ref{fig:re2}, we input the first frame ($T=0$) to three models, and all the models will output the reconstructed first frame, but CogVX-2b-nus outputs a misaligned frame, which also leads to more corrupted results in the subsequent frames. Vista and our model can reconstruct the first frame well. Although Vista has a high resolution of 576$\times$1024, its video content is not very clear, even a little blurred, which led to some subsequent corrupted results. In contrast, our model does not experience the accumulation of errors, and subsequent frames do not appear to be corrupted. The comparison validates the advantages of our method.

\paragraph{Long-term rollout}The proposed world model learns from the history sequence states and predicts the future sequence states, which enables it to plan and forecast the future. One of the strengths of our world model is the ability to make long-term future predictions. Two long-term future predictions are shown in Figure~\ref{fig:long}. Our model can predit diverse futures (such as slowdowns and deviations to the left), and the long-term videos' quality is well-maintained.

\begin{figure*}
    \centering
    \includegraphics[width=0.7\linewidth]{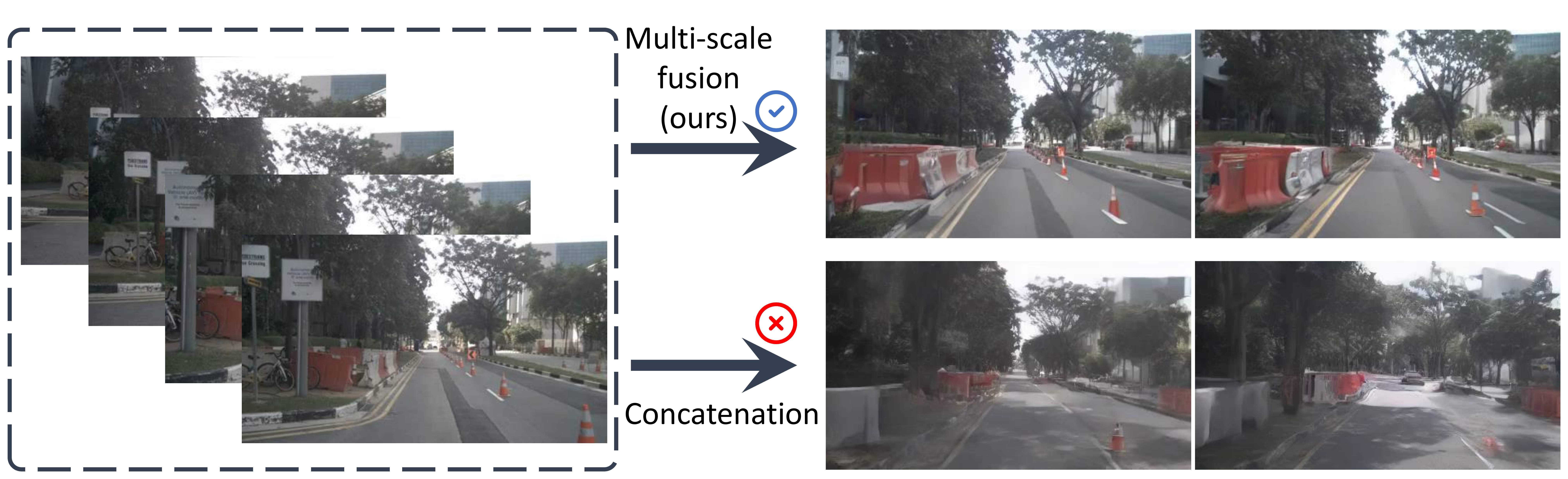}
    \caption{Comparison between the proposed multi-scale fusion and widely-used concatenation~\cite{yang2024cogvideox,blattmann2023svd}. Our method can converge faster and improve the quality of prediction.}
    \label{fig:ab1}
\end{figure*}

\subsection{Quantitative Evaluation}
\paragraph{Driving video generation}We compare with the state-of-the-art methods that can generate single-view driving videos. The results of ADriver-I~\cite{jia2023adriver} are not included since it generates 4 frames based on the previous 4 frames, which is unfair for the listed methods. GenAD~\cite{yang2024genad} and Vista~\cite{gao2024vista} results are also not included here, because they trained their models on extra large-scale video datasets. As shown in Table~\ref{tab:fvd}, even without the fine-grained conditions such as HDmap and 3Dbox as input priors, our method achieves the best results compared to other methods, with an FID of 6.9 and an FVD of 190.5, showing significant improvements.

\paragraph{Driving action prediction}We compare the driving action prediction with the previous work that has action prediction ability, including DriveDreamer~\cite{wang2023drivedreamer} and ADriver-I~\cite{jia2023adriver}. As shown in Tab.~\ref{tab:action}, our method achieves the lowest L1 error in speed, outperforming both DriveDreamer and ADriver-I. For the steering angle, our method also surpasses ADriver-I when using only actions, and comparable with ADriver-I using multi-modal inputs. Because all the methods finally use an MLP to project high-dimensional features into two low-level signals, we can directly report the average L1 error, and our method achieves the best average L1 error.

\subsection{Ablation Study}
\paragraph{Multi-scale fusion vs. Concatenation}We design a multi-scale fusion module to construct a state context short pathway. We compare a widely-used method (concatenation), which concatenates the context with the noisy latent features on the input layer~\cite{yang2024cogvideox,blattmann2023svd}. We modify the channels of $\texttt{conv\_in}$ to 8, to adapt the concatenated features. When training at the same steps, the model equipped with multi-scale fusion predicts better video than with concatenation in a large margin. The visualization results are in Figure~\ref{fig:ab1}.

\paragraph{Benefits of joint training}We ablate the different training strategies for video prediction, including no latent actions, frozen action module, and joint training. As shown in Table~\ref{tab:abaction}, joint training with the latent actions shows the best video quality, validating the effectiveness of our model.

\paragraph{Different reference frame}
Our multi-scale fusion module can tackle different reference frames, so we investigate the impact of the number of reference frames. As shown in Table~\ref{tab:ab2}, more frames can prompt the video consistency, i.e., lower FVD, and the impact on the image quality of each frame is comparable.
\begin{table}[]
   \caption{Video quality comparison under different training}
  \label{tab:abaction}
  \centering
\begin{tabular}{cccc}
\toprule
    Latent action & Joint train. & FID↓   & FVD↓      \\ 
    \midrule
     &  & 7.2 & 265 \\
    \checkmark &  & 7.1 & 227 \\
    \checkmark & \checkmark & 6.9 & 191 \\
    \bottomrule
  \end{tabular}
\end{table}

\begin{table}[]
   \caption{Video quality comparison using different reference frames in training}
  \label{tab:ab2}
  \centering
\begin{tabular}{lcc}
\toprule
    Frame Num. & FID↓   & FVD↓      \\ 
    \midrule
    1 & 6.9 & 190.5 \\
    2 & 6.9 & 188.3 \\
    4 & 7.1 & 182.5 \\
    \bottomrule
  \end{tabular}
  \vspace{-0.3cm}
\end{table}

\paragraph{Different resolution} We investigate the impact of training resolution on some metrics such as FID and FVD. As shown in Table~\ref{tab:ab3}, higher resolution can promote the FID, because the clarity of each frame is better. But higher resolution leads to worse FVD. One reason may be that dynamic information and motion at high resolution are difficult to learn, so the temporal consistency becomes worse.

\begin{table}[]
   \caption{Video quality comparison using different resolution.}
  \label{tab:ab3}
  \centering
\begin{tabular}{lcc}
\toprule
    Train. resolution & FID↓   & FVD↓      \\ 
    \midrule
    192$\times$384 & 6.9 & 190.5 \\
    256$\times$448 & 6.2 & 238.0 \\
    \bottomrule
  \end{tabular}
\end{table}
\section{Conclusion}
we proposed ProphetDWM, a novel end-to-end driving world model that can jointly predict future videos and actions. We construct a world model that is trained by learning latent actions given finite states and predicting actions and videos, where two crucial modules, the action module and diffusion-model-based transition model, are jointly optimized. Experiments in video generation and action prediction validate the effectiveness of our world model.



\clearpage
\setcounter{page}{1}
\maketitlesupplementary

\appendix

\renewcommand{\thetable}{A\arabic{table}}
\renewcommand{\thefigure}{A\arabic{figure}}

\section{Experiment details}
\label{sec:detail}
\begin{table}[]
    \centering
    \renewcommand{\arraystretch}{1.5}
    \begin{tabular}{lccc}
        \hline
        \textbf{Setting} & \textbf{1} & \textbf{2} & \textbf{3} \\ \hline
        \textbf{Resolution} & 192$\times$384 & 256$\times$448 & 256$\times$448 \\ \hline
        \textbf{Refer. Frames} & 1 & 1 & 4 \\ \hline
        \textbf{Sequence Length} & 8 & 10 & 10 \\ \hline
        \textbf{Fps} & 2  &  4  &  4 \\ \hline
        \textbf{LR}  & 1e-4  & 1e-4  & 1e-4 \\ \hline
    \end{tabular}
    \caption{Driving World Model experimental settings}
    \label{tab:detail}
\end{table}

In this paper, we conduct various experiment results. All experiments are conducted on A100 GPUs. The settings are shown in Table~\ref{tab:detail}. 

Following the previous setting~\cite{wang2023driveWM}, we conduct the first setting-1 for a fair quantitative comparison. The resolution is 192$\times$ 384, the sequence length is 8, the reference frame number is 1, and for simplicity, we do not utilize additional HDmap or 3Dbox as ~\cite{wang2023driveWM} did. The training dataset only includes keyframes, i.e., fps=2. In this setting, we report the FID and FVD results.

\begin{figure*}
    \centering
    \includegraphics[width=\linewidth]{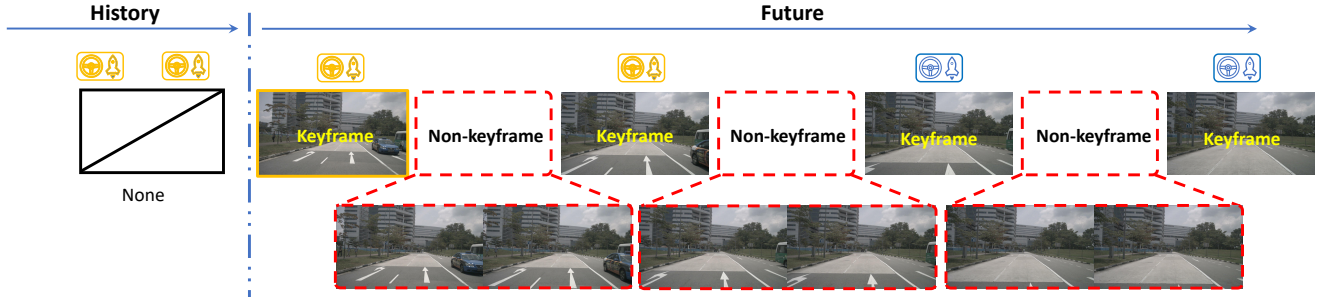}
    \caption{Training sequence visualization. The \textcolor{orange}{orange} marks are the inputs to our world model. The known future actions indicate the dynamic information in the future. Our model jointly predicts nine subsequent frames and two actions. }
    \label{fig:A3}
\end{figure*}

For the qualitative comparison, we train two models based on resolution 256$\times$448 and use one reference frame and four reference frames, respectively. The sequence length is 10. The training sequence is visualized in Figure~\ref{fig:A3}. We use 4Hz data, and there are two non-keyframes between two keyframes, and fps=4. The orange marks are the inputs to our world model. The known future actions indicate the dynamic information in the future. Our model jointly predicts nine subsequent frames and two actions. The reason is that in Nuscenes~\cite{caesar2020nuscenes}, the actions are only recorded for keyframes.

When training the image model, we finetune Stable-diffusion v1.4 on Nuscenes~\cite{caesar2020nuscenes} images for 50 epochs. When training the video model, we use a total batch size of 16 and train the video model for 30k steps in total. The learning rate is 1e-4, and the scheduler is a cosine scheduler. The video model's training took 40 hours on 4 A100 GPUs.

When inferring future videos and actions, we start from the reference frame(s) and reference actions, and the reference frame(s) and reference actions will be updated when a rollout is done. We roll out the future 16 times to produce a long-term future, with more than 100 frames.

\section{Quality comparison}

\begin{figure*}
    \centering
    \includegraphics[width=\linewidth]{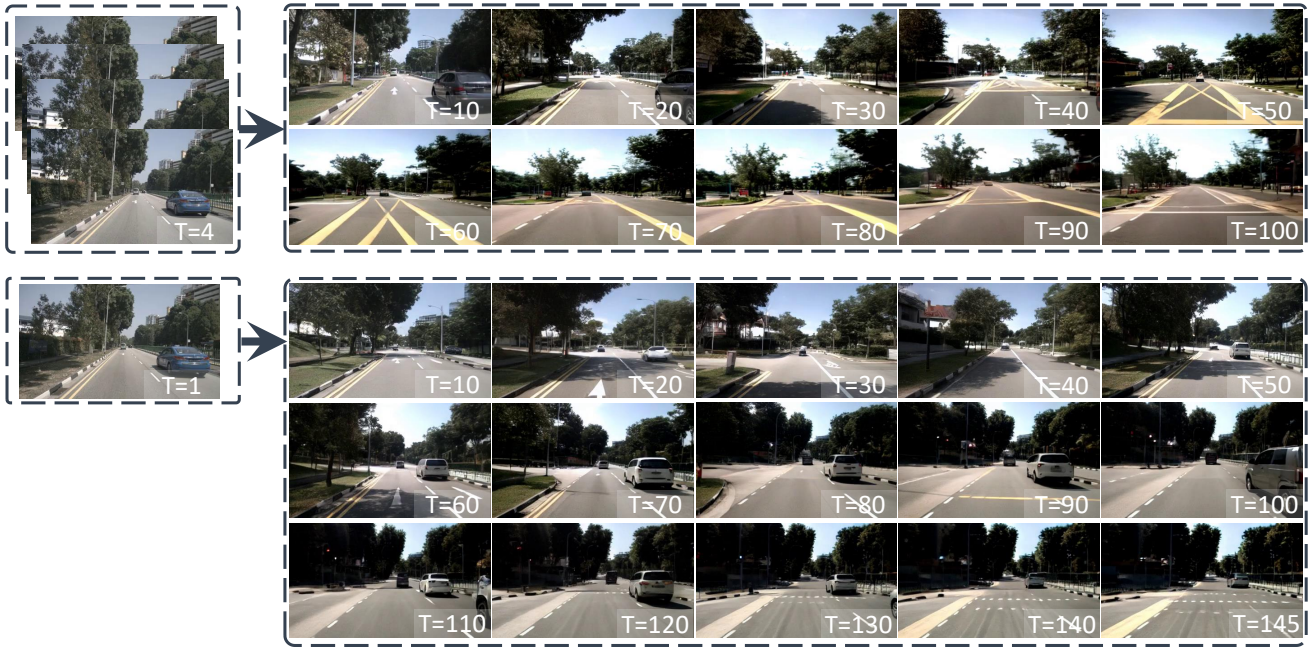}
    \caption{The comparison between using four reference frames and using one reference frame. Using the same 16 rollout times. Using four frames produces a total of 100 frames, and using one frame produces a total of 145 frames. Starting from the same scene, the results of different generations are still diverse.}
    \label{fig:A1}
\end{figure*}

We found that using more references would increase the video quality. As shown in Figure~\ref{fig:A1}, using more reference frames enhances content consistency and avoids excessive content changes. However, using only one frame produces more frames when using the same rollout times. For example, if we roll out the future 16 times, using four frames can produce 100 frames, while using one frame can produce 145 frames. As shown in Figure~\ref{fig:A1}, although these two predictions start from the same scene, the results of different predictions are still diverse.

\section{Action Control}
\begin{figure*}
    \centering
    \includegraphics[width=\linewidth]{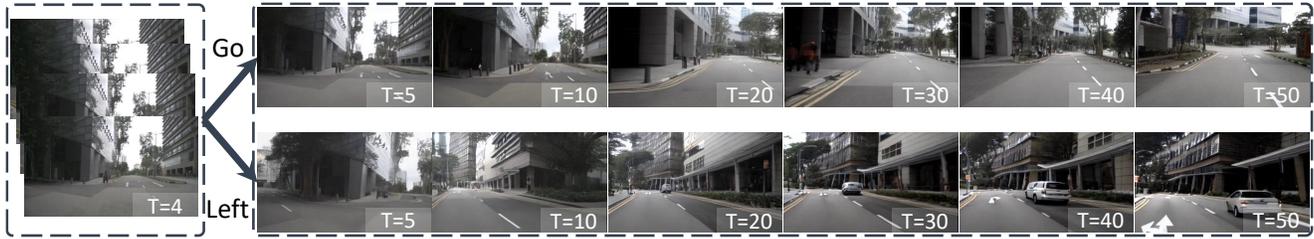}
    \caption{The prediction results from different low-level control actions.}
    \label{fig:A4}
\end{figure*}

In Figure~\ref{fig:A4}, we show the prediction results from different low-level control actions, such as going straight and turning left. We use low-level actions including speeds and steering angles to execute the commands. The results show that different commands can predict different futures while maintaining a high quality of the predicted results.

\section{More results}
\begin{figure*}
    \centering
    \includegraphics[width=\linewidth]{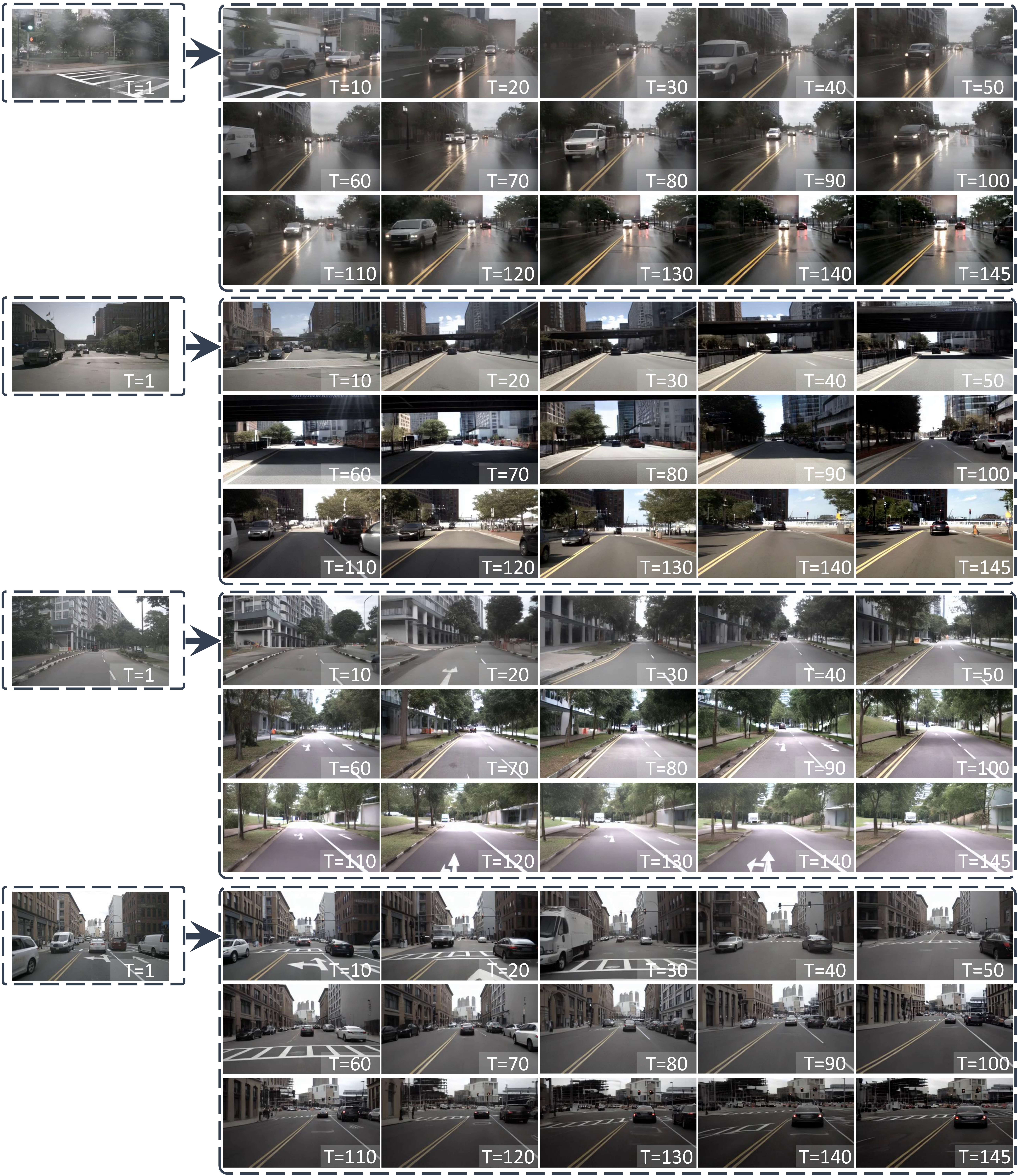}
    \caption{More results using one reference frame.}
    \label{fig:A2}
\end{figure*}

We present more results using one frame in Figure~\ref{fig:A2}. For each sample, we roll out 16 times to produce 145 frames.

{
    \small
    \bibliographystyle{ieeenat_fullname}
    \bibliography{main}
}


\end{document}